\documentclass[pdflatex,sn-nature]{sn-jnl}% Math and Physical Sciences Numbered Reference Style
\usepackage{graphicx}%
\usepackage{multirow}%
\usepackage{amsmath,amssymb,amsfonts}%
\usepackage{amsthm}%
\usepackage{mathrsfs}%
\usepackage[title]{appendix}%
\usepackage{xcolor}%
\usepackage{textcomp}%
\usepackage{manyfoot}%
\usepackage{booktabs}%
\usepackage{algorithm}%
\usepackage{algorithmicx}%
\usepackage{algpseudocode}%
\usepackage{listings}%
\usepackage{subcaption}
\usepackage{array}

\theoremstyle{thmstyleone}%
\theoremstyle{thmstyletwo}%

\theoremstyle{thmstylethree}%

\begin{document}

\title[Article Title]{What Makes Linguistic Representations Good Models of High-Level Visual Perception in the Human Brain?}


\author*[1]{\fnm{Anna} \sur{Bavaresco}}\email{a.bavaresco@uva.nl}

\author[1]{\fnm{Ina} \sur{Klarić}}\email{ina.klaric@student.uva.nl}

\author[1,2]{\fnm{Raquel} \sur{Fern\'andez}}\email{raquel.fernandez@uva.nl}\equalcont{These authors contributed equally to this work.}

\author[3]{\fnm{Marie-Francine} \sur{Moens}}\email{sien.moens@kuleuven.be}\equalcont{These authors contributed equally to this work.}

\affil[1]{\orgdiv{Institute for Logic, Language and Computation}, \orgname{University of Amsterdam}, \orgaddress{\street{Science Park 900}, \city{Amsterdam}, \postcode{1098XH}, \country{Netherlands}}}

\affil[2]{\orgdiv{European Commission, Joint Research Centre (JRC)}, \orgaddress{Brussels, Belgium}}

\affil[3]{\orgdiv{Faculty of Engineering Science}, \orgname{KU Leuven}, \orgaddress{\street{Celestijnenlaan 200A}, \city{Leuven}, \postcode{3001}, \country{Belgium}}}


\abstract{
Image descriptions represented with language models (LMs) predict human brain responses to naturalistic images in high-level visual regions, but the factors driving this predictivity remain unclear.  To investigate this, we systematically studied how images are described and which language models are used to embed those descriptions. For a common set of images, we considered six caption types---including human-annotated and multiple machine-generated captions---differing along several dimensions. Each caption was represented with five LMs, spanning autoregressive LMs trained to predict upcoming words and text embedders, i.e., LMs fine-tuned on semantic tasks requiring sentence/document-level representations. Machine-generated captions yielded significant brain predictivity and alignment, often surpassing human-annotated captions used in previous work. Across caption types, text embedders consistently outperformed autoregressive LMs, a pattern replicated when measuring behavioural alignment with image-similarity judgments.  Analyses of caption representations from different model layers further revealed that both brain predictivity and behavioural alignment peak at intermediate network depth, shortly after a point thought to mark the emergence of syntactic and semantic structure. Altogether, our results demonstrate that both the content of image captions and the LM used to represent them influence brain- and behaviour-modelling performance, establishing caption embeddings as a useful tool for studying high-level visual perception.  
}




\maketitle

% =============================================================
\section{Introduction}
\label{sec:introduction}
% =============================================================

Recent artificial intelligence (AI) models excel at a variety of tasks, while also providing new tools for cognitive (neuro)scientists to study the complex dynamics of the human brain and behaviour  \cite{serre2019deep, richards2019deep, kanwisher2023using, doerig2023neuroconnectionist, sucholutsky2025getting}.
In the visual perception domain, numerous studies have uncovered correspondences and differences between brain activations from visual areas and image features extracted from different computer-vision architectures \cite{kriegeskorte2015deep, yamins2014performance, eickenberg2017seeing}.

Early works revealed converging hierarchical processing in convolutional neural networks (CNNs) and the human visual system, showing that both extract low-level visual features (such as oriented bars and spatial frequencies) early and higher-level semantic properties later \cite{khaligh2014deep, yamins2014performance, gucclu2015deep}.
Studies adopting a behavioural perspective additionally found that image representations from late CNN layers align with similarity \cite{groen2018distinct, peterson2018evaluating}, typicality \cite{lake2015deep}, memorability \cite{dubey2015makes}, and shape \cite{kubilius2016deep} judgments, although humans rely on shape to a much larger extent than CNNs \cite{baker2018deep, geirhos2018imagenet}.

More recent research comparing CNNs against transformer-based architectures \cite{vaswani2017attention} and assessing the impact of various training regimes found that, despite known differences between representations learnt by CNNs and vision transformers \cite{raghu2021}, they exhibit similar brain predictivity, and that the training diet matters more than the learning objective \cite{conwell2024large}. Moreover, there is preliminary evidence that transformers also exhibit brain-like hierarchical processing of visual stimuli \cite{raugel2025disentangling}.

An initial explanation for the brain predictivity of artificial neural networks was that systems (computer-vision models and the human brain) optimised to solve a similar problem (e.g., identifying objects) are likely to learn similar image representations \cite{yamins2014performance, khaligh2014deep, zeman2020orthogonal}. This idea, already undermined by subsequent studies \cite{conwell2024large, bartnik2025representation}, has been further challenged by the recent finding that, in high-level visual areas, representations of image \textit{descriptions} (captions) extracted from large \textit{language} models (LLMs) trained solely on text predict brain responses to images as accurately as features extracted from vision models \cite{wang2023better, conwellrethinking, doerig2025high}. It is well-known that high-level visual areas selectively respond to semantically-coherent categories of visual stimuli, such as faces \cite{mccarthy1997face, kanwisher2006fusiform, tsao2006cortical}, hands \cite{bracci2010dissociable}, body parts \cite{downing2001cortical, peelen2005selectivity}, places \cite{epstein1998cortical, aguirre1998area}, and written words \cite{cohen2000visual, baker2007visual}. 
However, even admitting that LLM-derived caption representations are well-suited to capture semantic properties, their predictivity remains striking, as they lack information about strictly-visual image features. 

Despite attempts to determine whether the brain predictivity of image captions is driven by the semantics of the words they contain or the way in which these words are assembled within a sentence \cite{doerig2025high, conwellrethinking}, our understanding of the factors responsible for the brain predictivity of image captions remains preliminary. Moreover, all current results were obtained using captions from the MS COCO dataset \cite{lin2014microsoft} to predict fMRI activations from the Natural Scenes Dataset (NSD, \cite{{allen2022massive}}). This choice has practical advantages, as the NSD brain responses were elicited by images from MS COCO; at the same time, the quality of the captions provided in MS COCO may be suboptimal, as they are relatively short ($\sim11$ words) and provided by crowdworkers.

To broaden our understanding of the brain-modelling potential exhibited by caption representations, we assessed their brain predictivity and alignment by analysing existing fMRI responses in high-level visual areas, systematically studying the effect of two factors: the type of caption, and the language model (LM) used to derive caption representations. We obtained multiple captions for each image, varying in length and amount of detail, by generating them with vision-language models (VLMs); we next extracted caption representations using five different LMs trained solely on text. Our results show that machine-generated captions significantly predict brain activations, often outperforming the human-annotated captions used in previous works. Moreover, the LM used to represent captions can significantly impact their brain predictivity and alignment, with recent text embedders outperforming autoregressive LMs. This pattern also holds when measuring caption representations' alignment with human image-similarity judgments and, remarkably, caption representations align with those judgments better than image features computed with vision-only models. 

\begin{figure}
\centering
\begin{subfigure}[c]{0.98\textwidth}
    \centering
    \includegraphics[width=\linewidth]{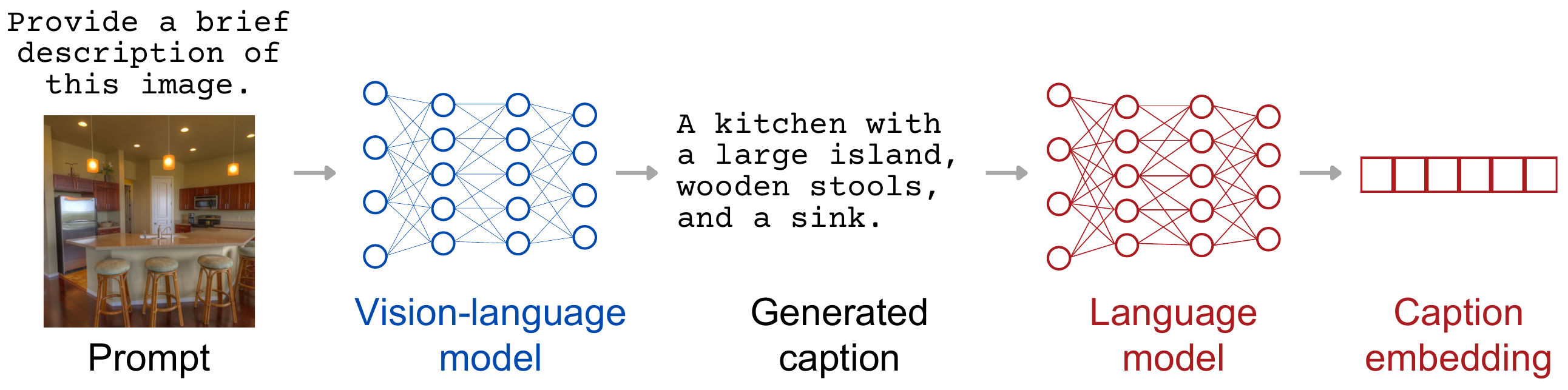}
    \caption{}
    \label{fig:captions_upper}

\end{subfigure}
\begin{subfigure}[c]{0.25\textwidth}
    \centering
    \includegraphics[width=\linewidth]{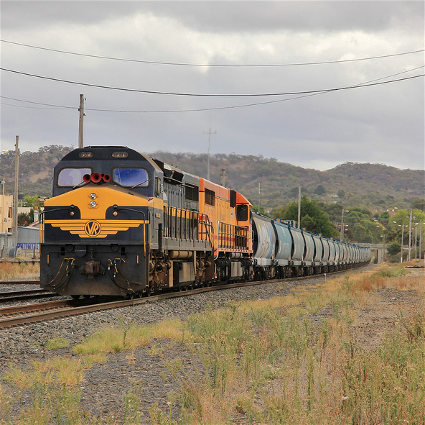}
    \caption{}
    \label{fig:captions_lower_left}
\end{subfigure}
\hfill
\begin{subfigure}[c]{0.7\textwidth}
    \centering
    \includegraphics[width=\linewidth]{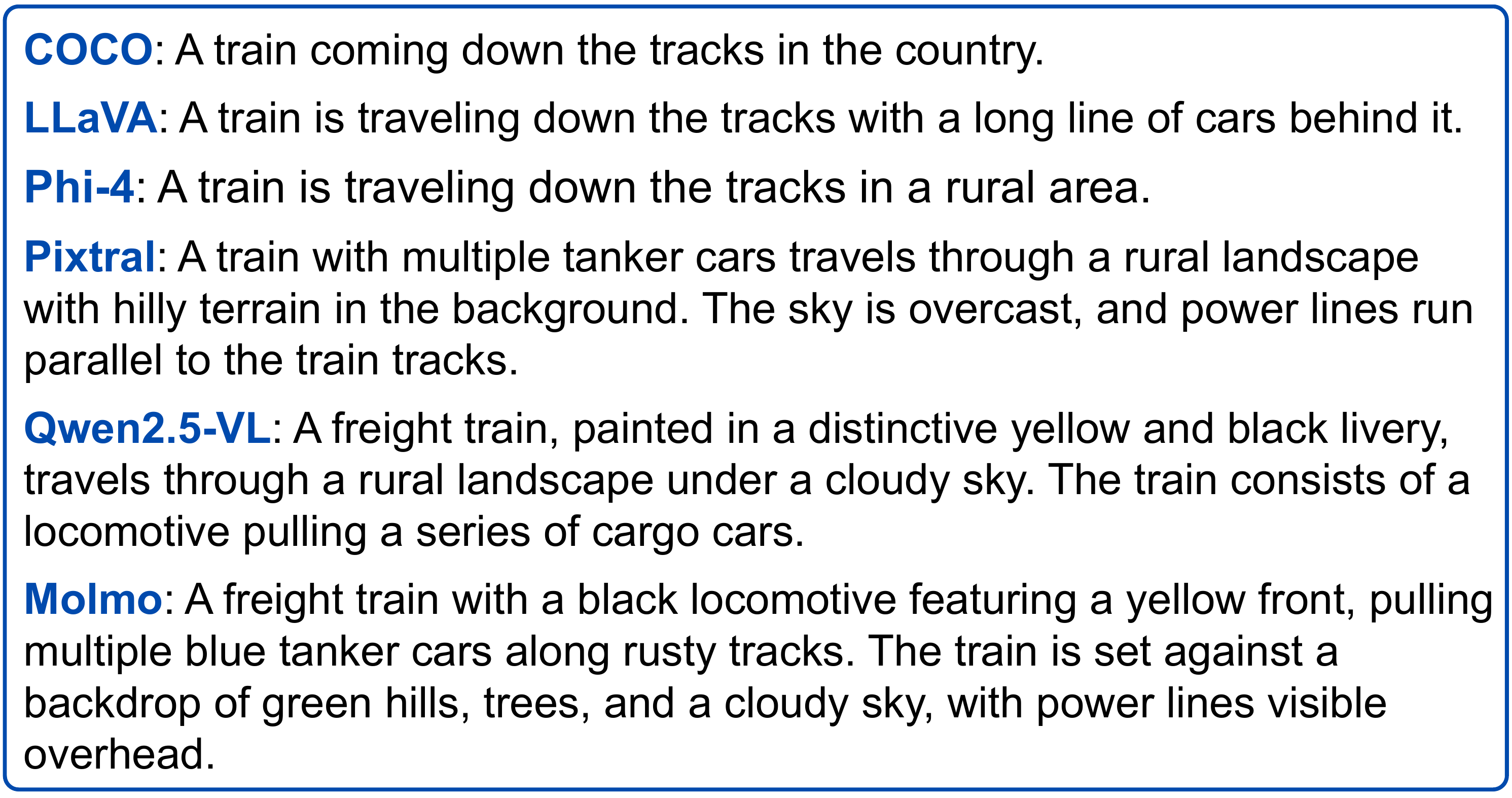}
    \caption{}
    \label{fig:captions_lower_right}

\end{subfigure}

\caption{\textbf{Overview of generated captions and their embedding.} The upper panel (\textbf{a}) shows a schematic of the pipeline used to create caption embeddings from machine-generated captions. Each vision-language model was prompted to create a brief description of the input images. The machine-generated captions were then embedded with a language model. In the lower panels, an example image (\textbf{b}) along with different captions (\textbf{c}), including one from MS COCO and five machine-generated.}
\label{fig:captions}
\end{figure}

 % =============================================================
\section{Results}
\label{sec:results}
% =============================================================
% [Should be $\sim$ 2900--3100 words, it's now about 2100]

We investigated the brain predictivity of caption embeddings by analysing the natural scenes dataset (NSD, \cite{allen2022massive}), which provides a large-scale collection of fMRI responses. These were recorded from participants ($n=8$) viewing naturalistic scene images from the MS COCO dataset \cite{lin2014microsoft} while performing a memory task. Crucially, no image descriptions were presented to participants at any point in the experiment. 

To ensure the generalisability of our findings across brains, we focused on the subset of 906 images viewed at least once by all participants. We considered three functionally-localised regions of interest (ROIs): \textit{face}-, \textit{body}-, and \textit{place}-selective; these are high-level visual areas where previous studies documented significant predictivity for caption embeddings \cite{wang2023better, doerig2025high}. Specifically, we aimed to predict the voxel-wise neural activations (beta coefficients) in these ROIs.

We assessed the brain predictivity achieved by the MS COCO captions used in previous work, provided by human crowdworkers, and by five machine-generated caption sets. These were obtained by instructing five open-source vision-language models (VLMs)---LLaVA OneVision \cite{li2025llavaonevision}, Phi-4 \cite{abouelenin2025phi}, Pixtral \cite{pixtral}, Qwen2.5-VL \cite{bai2025qwen2}, and Molmo \cite{deitke2024molmo}---to provide factual descriptions of the images, as shown in Fig.~\ref{fig:captions_upper}. The resulting caption sets vary systematically, as each VLM exhibits its specific `writing style' (see Fig.~\ref{fig:captions_lower_left} and~\ref{fig:captions_lower_right} for a qualitative example).

To quantitatively measure these differences across caption types, we computed the following metrics: length (average number of words), perplexity, visualness, and lexical density. Perplexity ($PPL$) is a standard metric in natural language processing (NLP), estimating how `surprising' a text sequence is to a language model (in our case, Ministral3 \cite{liu2026ministral}); it is commonly intended as a proxy for fluency (low perplexity indicates high fluency).
The visualness of a caption was computed by summing the visualness ratings of its constituent words, taken from the Lancaster Sensorimotor Norms \cite{lynott2020lancaster}. Finally, lexical density was calculated as the percentage of content words (nouns, adjectives, verbs, and adverbs) present in a caption. More details on how these metrics were computed are provided in Methods. The average values of all metrics per caption type are reported in Tab.~\ref{tab:caption_metrics}.

An inspection of these metrics reveals several crucial differences among caption types. First, even the least fluent machine-generated caption type (Phi-4, $PPL=49.64$) is significantly more fluent ($W=26759$, $p\ll0.001$) than the human-annotated MS COCO captions ($PPL=134.14$), as assessed with a Wilcoxon signed-rank test. In addition, three out of the six caption types are relatively short ($11<\#words<13$), while the remaining three are substantially longer ($28<\#words<37$). Caption length positively correlates with visualness for all caption types ($0.77<r<0.88$, $p\ll0.001$), confirming that longer captions provide additional visual information; they also exhibit higher adjective density ($12\%<LD_{adj}<16\%$) than shorter captions ($6\%<LD_{adj}<8\%$), suggesting that short captions simply mention entities, while long ones also describe them. Overall, these metrics show that caption types systematically differ in fluency and detail (reflected by length and differences in lexical densities), allowing us to check whether these aspects matter for brain predictivity. 

\begin{figure}
    \centering
    \includegraphics[width=0.75\linewidth]{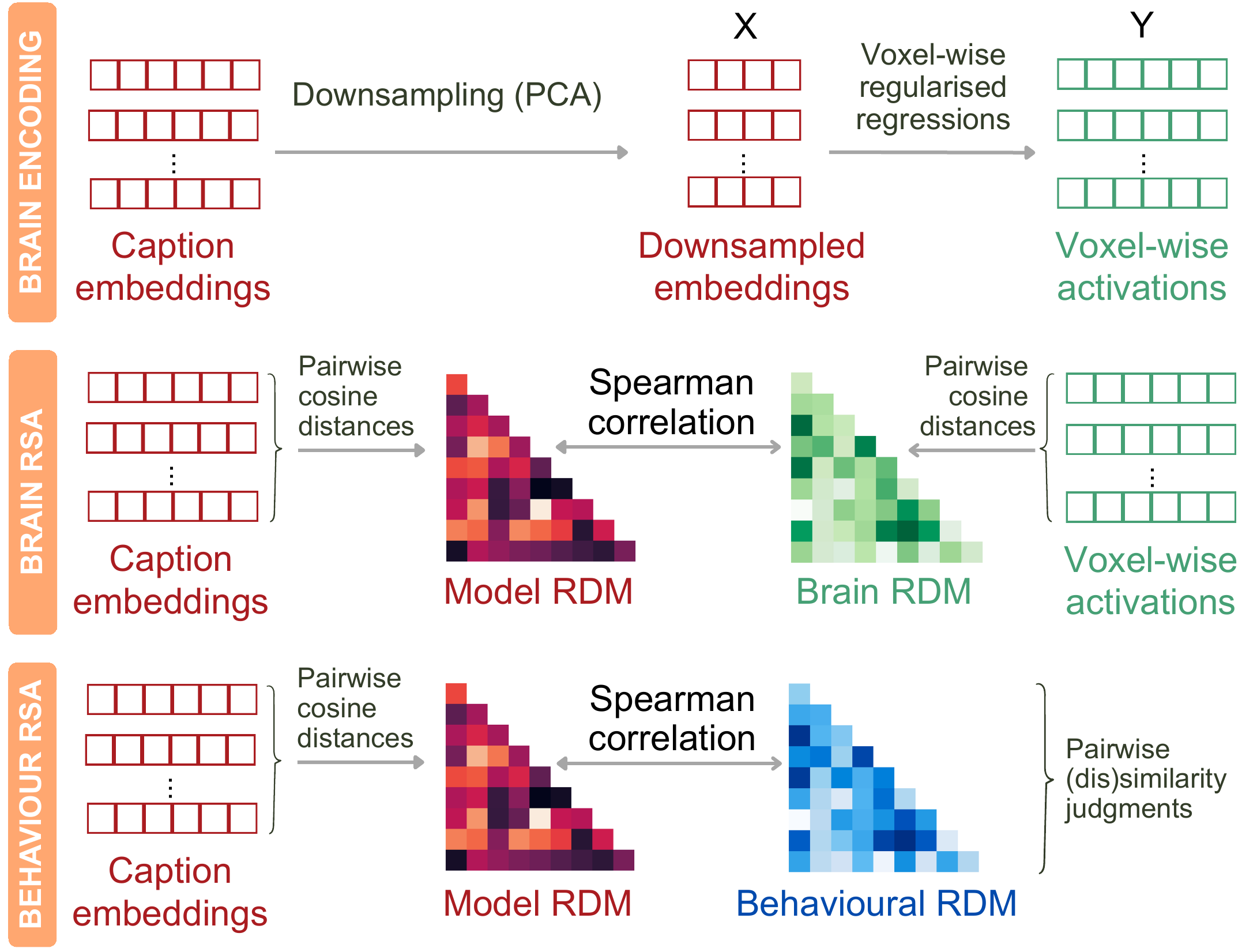}
    \caption{\textbf{Evaluation methods used to assess the brain- and behaviour-modelling properties of caption embeddings.} In brain encoding, caption embeddings are downsampled with PCA and subsequently used as dependent variables in Ridge regression models trained to predict voxel-wise activations. In representational similarity analysis (RSA), representational dissimilarity matrices (RDMs) are created by computing pairwise cosine distances between caption embeddings or neural activations. The alignment between representational spaces is then quantified by computing a Spearman correlation ($\rho$) between the vectorised off-diagonal elements of the RDMs. The same RSA procedure was used to quantify captions' alignment to behavioural measures---i.e., human (dis)similarity judgments---arranged in an RDM where entries corresponded to the perceived dissimilarity between image pairs.}
    \label{fig:evaluation_pipeline}
\end{figure}

\begin{table}[]
    \centering
    \begin{tabular}{llllllll} \toprule
      \textit{Caption type}   & \textit{\#words} & \textit{PPL} & \textit{Visualness} & $LD_{nn}$ & $LD_{adj}$ & $LD_{vb}$ & \textit{LD}\\\midrule
      MS COCO & 11 (2) & 134 (130) & 28	(7) & 35	(9) & 7	(8) & 10	(7) & 54	(10)\\
      LLaVA & 13 (3) & 39 (23) & 37 (8) & 34	(8) & 8	(8) & 8	(5) & 52	(7) \\
      Phi-4 & 11 (2) & 50 (32) & 29 (6) & 34	(7) & 6	(7) & 8	(6) & 48	(7) \\
      Pixtral & 31 (5) & 15	(6) & 74	(12) & 30	(5) & 	12	(6) & 11	(4) & 55	(6) \\
      Qwen2.5-VL & 28 (7) & 19 (9) & 70 (16) & 33	(6) & 12	(6) & 11	(4) & 58	(6) \\
      Molmo & 37 (8) & 13 (5) & 93 (18) & 31	(5) & 16	(5) & 10	(3) & 57	(5) \\ \bottomrule
          
    \end{tabular}
    \caption{\textbf{Differences across caption types quantified with several metrics.} The table reports average values across captions for each caption type, with standard deviations in brackets. The metrics are: caption length, expressed as number of words (\textit{\#words}); perplexity (\textit{PPL}); visualness, expressed as the sum of word-level visual ratings (ranging from 0 to 5, with 0 indicating `not experienced at all with vision'); lexical density (\textit{LD}), expressed as the percentage of content words (nouns, adjectives, verbs, adverbs). More specific density measures were additionally computed for nouns ($LD_{nn}$), adjectives ($LD_{adj}$), and verbs ($LD_{vb}$).}
    \label{tab:caption_metrics}
\end{table}

Besides evaluating the impact of caption types, we additionally asked whether brain predictivity is influenced by the language model (LM) used to \textit{embed} the captions.  In NLP, `embedding' usually means transforming text into vector representations (\textit{embeddings}). Depending on the model employed to compute the embeddings, the resulting representations vary, and so may their brain predictivity. We therefore embedded each caption with five different open-source, transformer-based LMs, all trained unimodally (only on text, not on images). We did not use VLMs trained on text and visual input (images or videos) to compute embeddings to avoid any `contamination' with visual information, as we aimed to assess the brain predictivity of \textit{linguistic} representations.

The LMs used to compute embeddings were: an early transformer trained with masked language modelling (BERT, \cite{devlin-etal-2019-bert}); two autoregressive LMs (GPT-2 \cite{radford2019language} and Llama3.1 \cite{grattafiori2024llama}) trained to predict upcoming tokens (chunks of text); and two recent text embedders extensively fine-tuned in a self-supervised fashion on datasets spanning semantics tasks, such as retrieval, classification, and semantic text similarity (Qwen3 Embedding \cite{zhang2025qwen3} and KaLM Embedding \cite{zhao2025KaLM}). These models cover architectures widely studied by both the NeuroAI and the NLP community (BERT and GPT-2), as well as more recent ones. Each LM was used to compute embeddings for all caption types; we analysed embeddings extracted from all model layers, as the last one is rarely the most brain-predictive \cite{schrimpf2021neural, caucheteux2022brains}.

 We additionally measured the brain predictivity achieved by image features computed with vision models, as they provide an informative reference. These features were obtained by feeding two widely-known computer-vision architectures, ResNet-50 \cite{he2016deep} and ViT \cite{dosovitskiyimage}, with the 906 images. Each model was considered in three variants differing in training regime: vision-only self-supervision (DINO \cite{caron2021emerging} or DINOv2 \cite{oquab2024dinov} frameworks), minimal language supervision (image labels from ImageNet \cite{deng2009imagenet}), and full language supervision (image captions, as per CLIP framework \cite{radford2021learning}).

\begin{figure}[h!]
    \centering
    \begin{tabular}{cc}
      \includegraphics[width=0.64\textwidth]{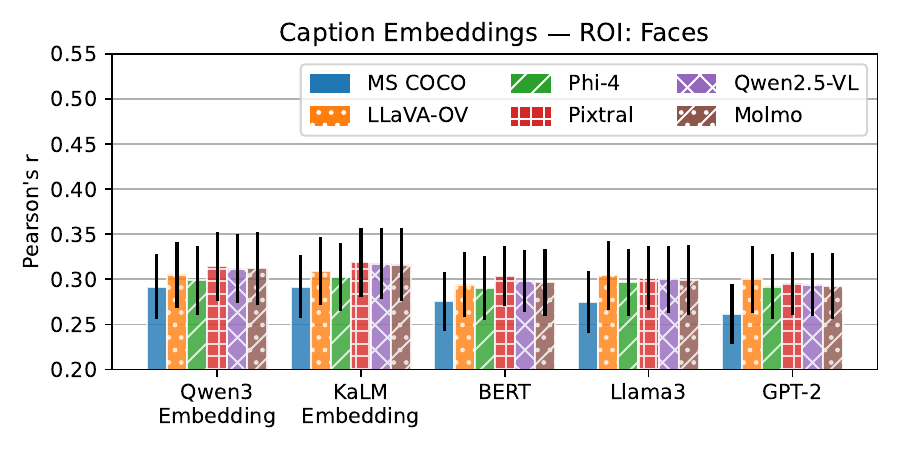}  & \includegraphics[width=0.33\textwidth]{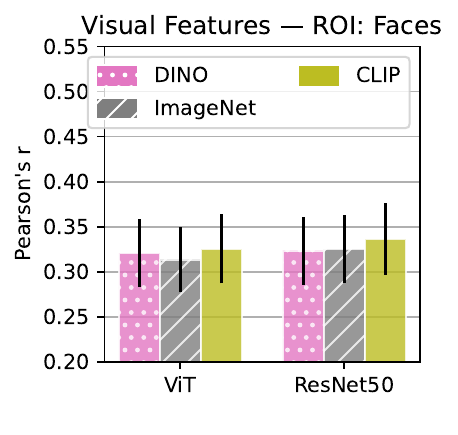}  \\
      \includegraphics[width=0.64\textwidth]{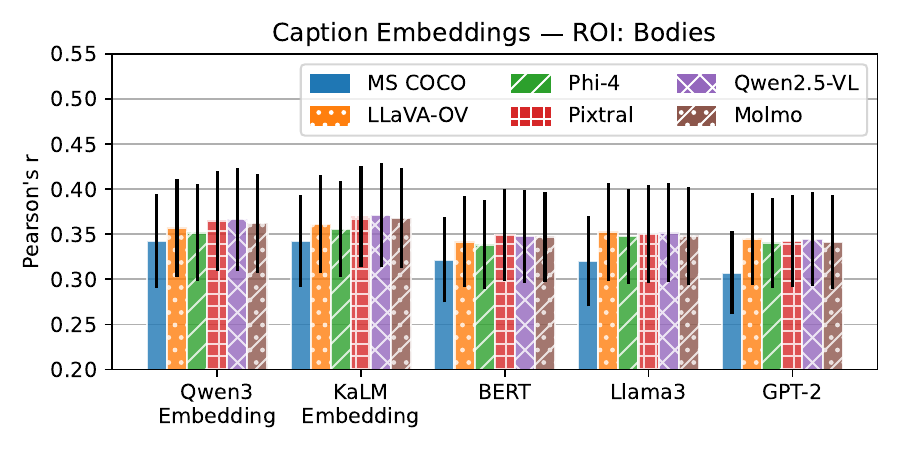}  & \includegraphics[width=0.33\textwidth]{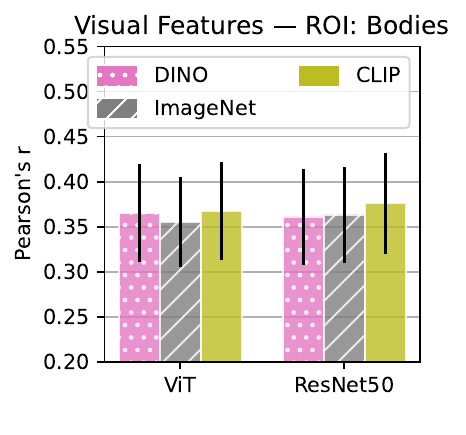}  \\
      \includegraphics[width=0.64\textwidth]{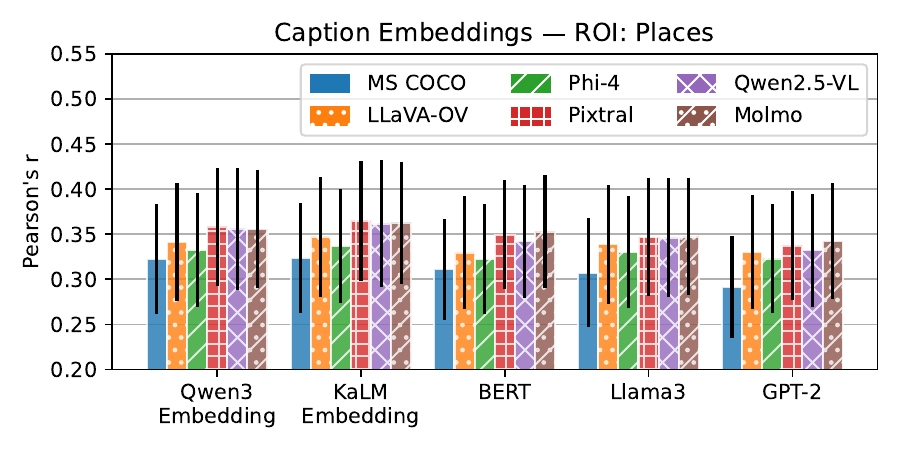}  & \includegraphics[width=0.33\textwidth]{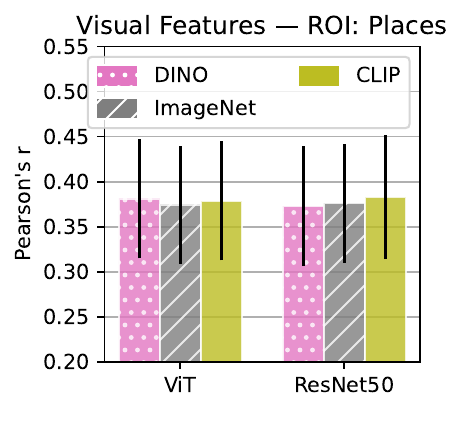}  \\
         
    \end{tabular}
    \caption{\textbf{Brain encoding results.} Brain encoding performance achieved by caption embeddings (left-hand panels) and image features (right-hand panels) in high-level visual ROIs, measured as Pearson's correlations ($r$) between predicted and observed neural activations, computed on held-out test sets resulting from a 5-fold cross-validation framework. Results are averaged across the voxels belonging to each ROI (face-, body-, and place-selective); for each voxel, we considered the correlation yielded by the most predictive model layer. Note the scale of the $y$ axis, starting from $0.2$. Additionally, the average was computed only on voxels where predictions were statistically significant, as assessed with permutation tests (1000 permutations, FDR-corrected $p$-values). Error bars indicate the standard deviation across participants.}
    \label{fig:enc_acc}
\end{figure}

\begin{figure}
    \centering
    \begin{tabular}{c}
         \includegraphics[width=0.85\textwidth]{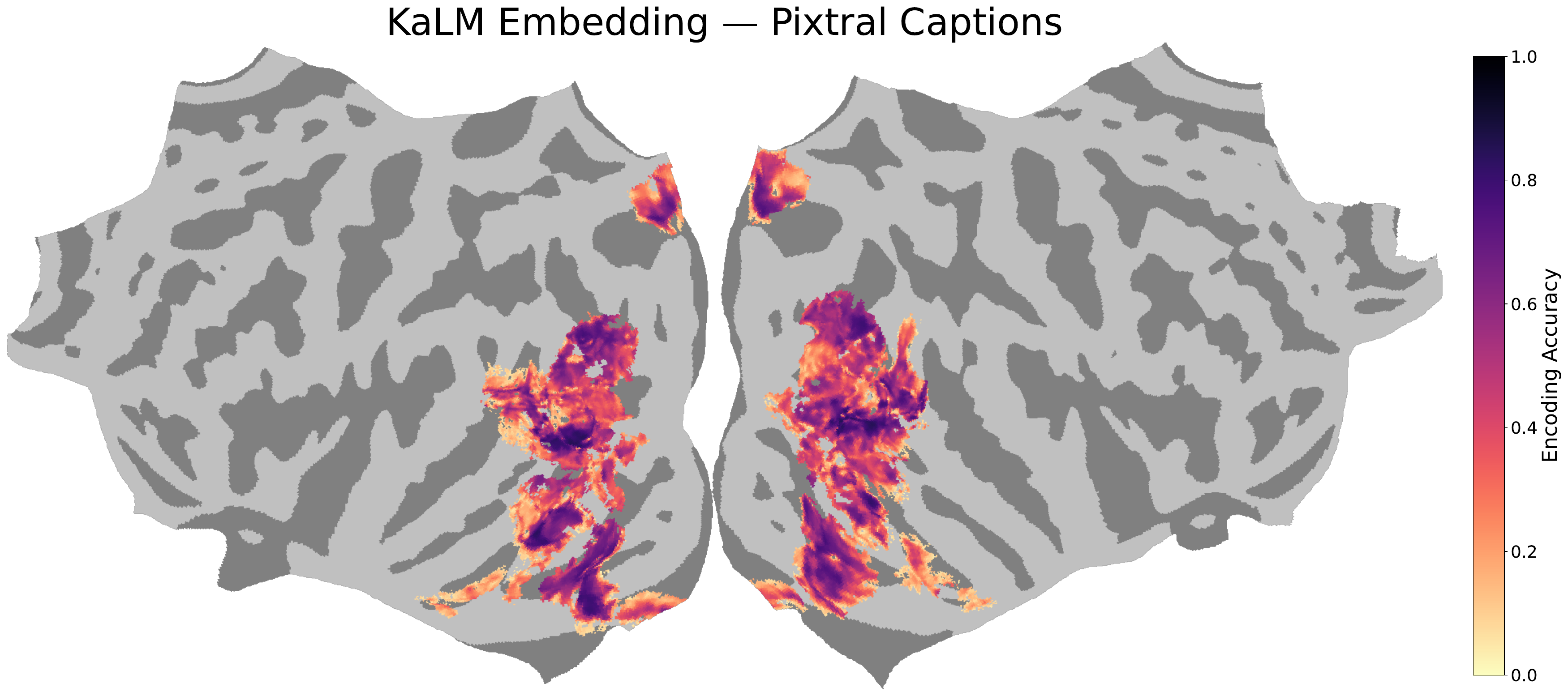}  \\
         \includegraphics[width=0.85\textwidth]{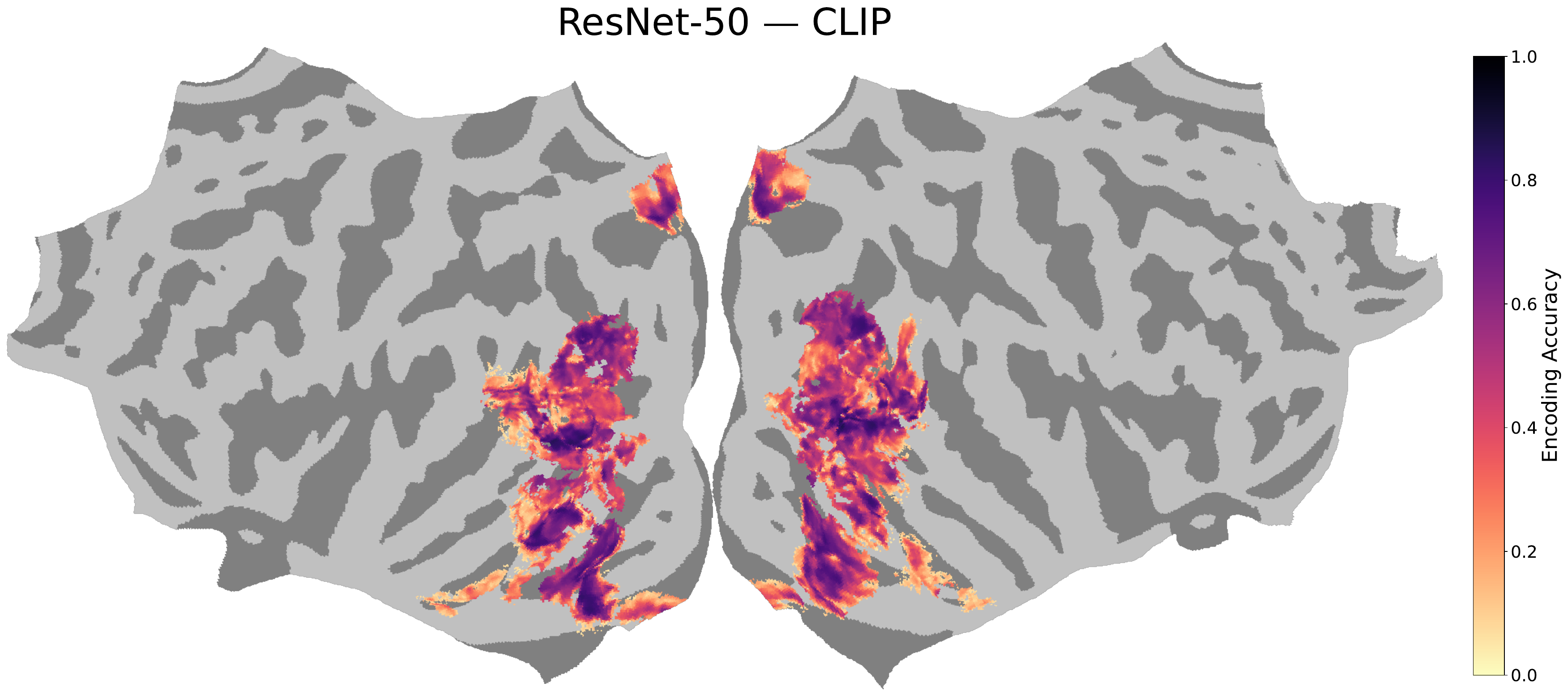} 
         
    \end{tabular}
    \caption{\textbf{Brain encoding performance in high-level visual cortices.} Brain flatmaps showing the accuracy of encoding models for the voxels from place-, face-, and body-selective areas that could be predicted significantly, as assessed with permutation tests (1000 permutations, FDR-corrected $p$-values). The top panel shows the results obtained by training the encoding models from Pixtral-generated captions embedded with KaLM Embedding. The lower panel shows the encoding accuracy obtained with image features extracted from a ResNet-50 model trained with the CLIP framework. Both flatmaps show the accuracy for participant 5, whose fMRI recordings were the least noisy. Vertices are mapped to the `fsaverage' space for visualisation purposes only; results were originally computed in the participant's native space. Predictivity patterns obtained from caption embeddings and visual features are strikingly similar.}
    \label{fig:brain_flatmaps}
\end{figure}

% ------------------------------------------------------------
\subsection{Machine-generated captions significantly predict neural activity in high-level visual areas}
\label{sec:caption_embedder_vs_caption_type}
% ------------------------------------------------------------
We assessed the brain predictivity of caption embeddings obtained from different LMs with \textit{voxel-wise encoding models} \cite{naselaris2011encoding, dupre2025voxelwise}. This approach consists in training regularised linear regressions to independently predict voxel-wise activations from caption embeddings or image features, as schematised in Fig.~\ref{fig:evaluation_pipeline}. The accuracy of the encoding models was evaluated as the Pearson correlation between the predicted and observed neural activations, computed on a test set. A high correlation suggests that the features making up caption embeddings may be similar to those employed by the brain to represent images in the target ROIs.  As we assessed with permutation tests, predictions were statistically significant for at least 90\% of the voxels in all cases, confirming that both caption embeddings and image features capture brain-relevant properties in high-level visual areas. Correlations were \textit{not} normalised by the amount of explainable variance, as doing so may conflate data quality and observed effect size \cite{lage2019methods}, but we documented systematically higher encoding performance in voxels with greater explainable variance (see Supp. Figs. 2--6).\footnote{All Supplementary Materials are provided in our public GitHub repository: \url{https://github.com/dmg-illc/linguistic-representations-visual-perception}}

As shown in Fig.~\ref{fig:enc_acc}, the best-performing caption embeddings are only marginally less brain predictive than image features in all ROIs. Moreover, the most predictive caption embeddings and image features exhibit similar predictivity patterns across high-level visual regions (see Fig.~\ref{fig:brain_flatmaps}). This confirms that caption embeddings can significantly predict neural activations in high-level visual areas even if they do not represent any strictly-visual property.

We next analysed brain-predictivity differences among caption types and LMs using linear mixed-effects models. We fit an independent model in each ROI, with \textit{participant} as a random effect, and \textit{caption type} and \textit{LM} as fixed effects; the dependent variable was the Pearson correlation between observed and predicted neural activations. MS COCO captions embedded with Qwen3 were set as the reference condition (Bodies: $\beta=0.34$, $SE=0.02$, $t=16.10$; Faces: $\beta=0.29$, $SE=0.13$, $t=20.67$; Places: $\beta=0.32$, $SE=0.02$, $t=13.25$). Coefficients obtained in this condition highlighted cross-ROI variations in average predictivity. However, these differences are likely attributable to varying amounts of noise (see Supp. Fig. 1); that is, predictivity was higher simply because noise was lower.

Considering caption types, the MS COCO ones proved to be the \textit{least} predictive, while those generated by Qwen2.5-VL exhibited the highest predictivity in body-selective areas ($\beta=0.37$, $SE=0.00$, $t=25.54$), those generated by Pixtral in face-selective areas ($\beta=0.31$, $SE=0.00$, $t=23.61$), and those generated by Molmo in place-selective areas ($\beta=0.36$, $SE=0.00$, $t=28.10$). This finding has two important implications. First, it shows that machine-generated captions can significantly predict neural activations. Second, it confirms that the previously highlighted limitations of MS COCO captions are substantiated. 

Indeed, we observed the highest brain predictivity for caption types (Qwen2.5-VL-, Pixtral-, and Molmo-generated) containing significantly more visual information than the MS COCO ones, as suggested by the visualness and lexical density metrics we computed (see Tab.~\ref{tab:caption_metrics}). Interestingly, even captions by Phi-4, which are quite short and almost as visual as the MS COCO ones, marginally outperformed them; this indicates that higher caption fluency quantified as LLM perplexity---that did differ substantially between Phi-4 ($PPL=49.64.14$) and MS COCO captions ($PPL=134.14$, $W = 26759$, $p \ll 0.001$)---is associated with increased brain predictivity.

Regarding the LMs used to derive the caption embeddings, analyses with mixed-effects models suggest that Qwen3 and KaLM Embedding have an advantage; indeed, only captions embedded by KaLM outperformed the reference level Qwen3 across all ROIs (Bodies: $\beta=0.34$, $SE=0.00$, $t=3.48$; Faces: $\beta=0.29$, $SE=0.00$, $t=3.36$; Places: $\beta=0.32$, $SE=0.0$, $t=3.91$). 
This means that recent text embedders (Qwen3 and KaLM Embedding) are not only superior to earlier LMs (BERT) and autoregressive (L)LMs (GPT-2 and the more recent Llama3.1) on text-embedding benchmarks, but also yield text representations that better capture brain-relevant information. More broadly, our results indicate that the LM used to represent the captions does play a crucial role in enhancing their brain-relevant properties.

\begin{figure}[h!]
    \centering
    \begin{tabular}{cc}
      \includegraphics[width=0.62\textwidth]{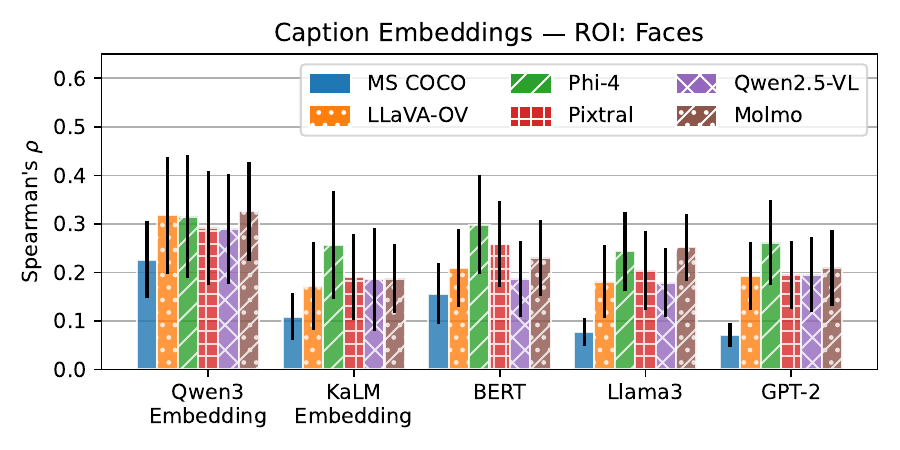}  & \includegraphics[width=0.31\textwidth]{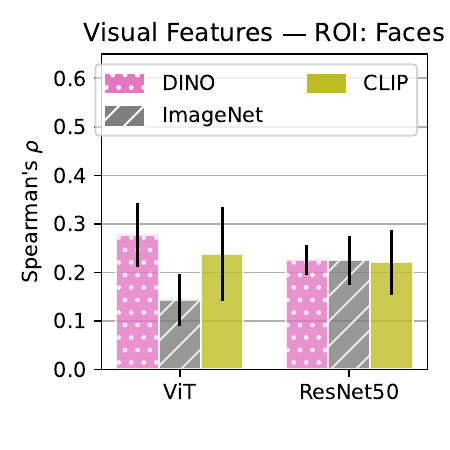}  \\
      \includegraphics[width=0.62\textwidth]{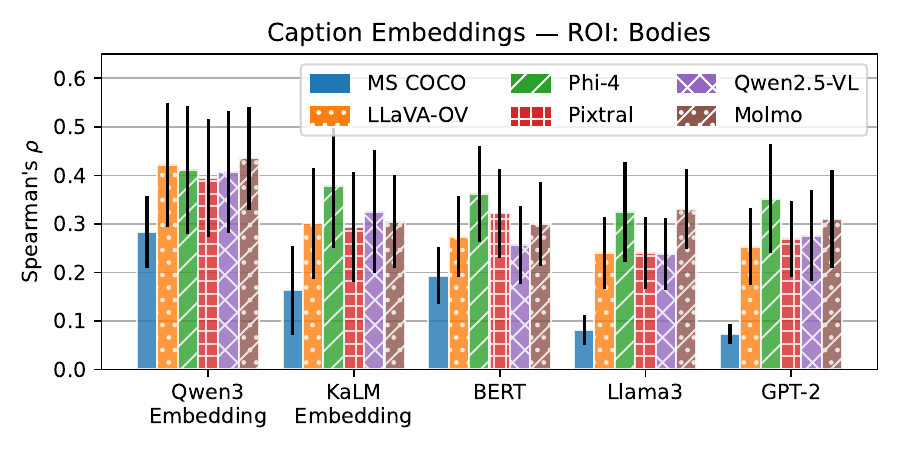}  & \includegraphics[width=0.31\textwidth]{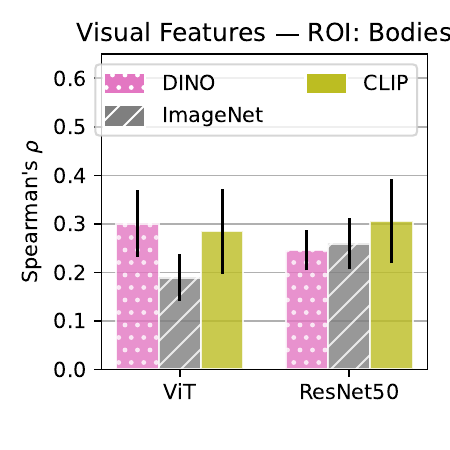}  \\
      \includegraphics[width=0.62\textwidth]{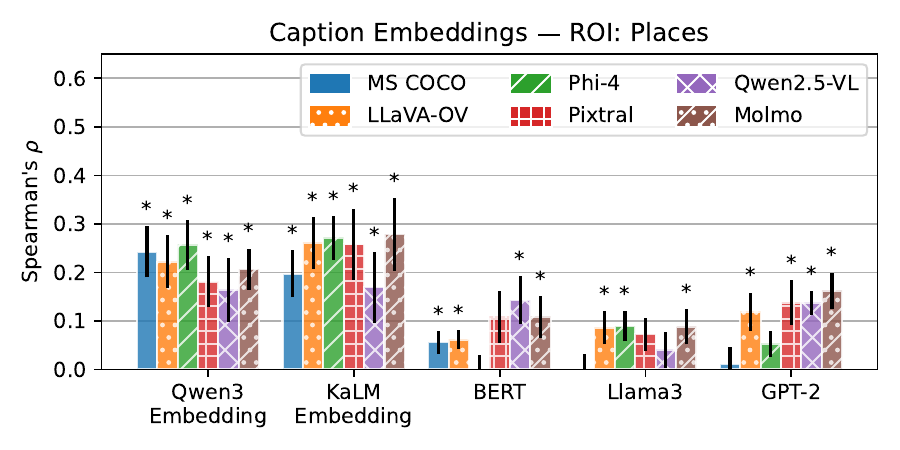}  & \includegraphics[width=0.31\textwidth]{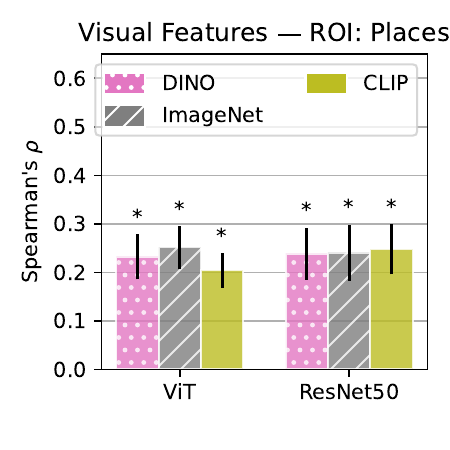}  \\
         
    \end{tabular}
    \caption{\textbf{Brain alignment achieved by caption embeddings.} Results of representational similarity analysis computed between caption embeddings (left-hand panels) or image features (right-hand panels) and neural activations. Spearman correlations ($\rho$) indicate the strength of the alignment between the two representational spaces, averaged across participants; error bars indicate the standard deviation across participants. Correlations are displayed only for the most aligned model layer, which was selected participant-wise. Alignment is displayed separately for each brain ROI: in the upper panel for face-selective regions, in the mid panel for body regions, and in the lower panel for place-selective areas. To facilitate comparisons across ROIs, all results have been normalised by the noise ceilings. All correlations are statistically significant in \textit{Faces} and \textit{Bodies}, as assessed with permutation tests (1000 permutations, $p<0.001$). In \textit{Places}, asterisks indicate statistically significant correlations across all participants.}
    \label{fig:brain_rsa}
\end{figure}

% ------------------------------------------------------------
\subsection{Caption representational spaces are brain-aligned, with differences across ROIs}
\label{sec:brain_rsa}
% ------------------------------------------------------------
 We next computed \textit{brain alignment}, which measures correspondences between brain activations and caption embeddings, providing information complementary to brain encoding. While brain encoding assesses predictivity at the feature level, alignment measures correspondences at the level of representational spaces---it quantifies the extent to which images yielding similar neural activations are also similar in the caption-embedding space. We computed brain alignment using a standard methodology: Representational Similarity Analysis (RSA, \cite{kriegeskorte2008representational}), illustrated in Fig.~\ref{fig:evaluation_pipeline}.

Results from RSA revealed that all caption embeddings and image features yield statistically significant alignment ($p<0.001$) in face- and body-selective ROIs. In place-selective regions, we observed some non-statistically-significant correlations; however, this never happened for more than three participants. Correlations remained substantially lower than the noise ceilings in all ROIs, especially in the place-selective ROI, indicating that caption embeddings and image features account for less than half of the explainable variance in the similarity patterns of fMRI activations. 

As shown in Fig.~\ref{fig:brain_rsa}, the most aligned caption embeddings approximate similarity patterns of brain activations as accurately as image features from vision models. This means that, while the \textit{features} captured by caption embeddings are slightly less brain predictive than image features (see Fig.~\ref{fig:enc_acc}), the representational geometry they define is comparably brain-aligned, likely due to similar co-occurrence patterns between images and their linguistic descriptions.

As for brain encoding, we analysed alignment differences across caption embeddings with linear mixed-effects models, fit separately for each ROI. The variables and reference conditions considered were the same as above, with the only difference that we included an interaction term between \textit{caption type} and \textit{LM}, as this helped explain additional variance. The estimated intercepts in the reference condition---MS COCO captions embedded with Qwen3---were $\beta=0.20$ ($SE=0.00$; $t=7.78$) in body-selective areas, $\beta=0.12$ ($SE=0.02$; $t=7.00$) in face-selective areas, and $\beta=0.14$ ($SE=0.01$; $t=14.00$) in place-selective areas. In this case, differences among ROIs cannot be attributed to differing amounts of noise in the fMRI recordings. Indeed, as visible in Fig.~\ref{fig:brain_rsa}, where correlations are normalised by the noise ceilings, caption embeddings explain significantly more variance in body and face areas than in place-selective ones. 

In the body-selective ROI, MS COCO captions embedded with Qwen3 were significantly outperformed ($p\ll0.001$) by all machine-generated caption types embedded with the same model, with Molmo-generated captions resulting in the highest coefficient estimates ($\beta=0.24$, $SE=0.12$). Moreover, embedding MS COCO captions with LMs different from Qwen3 resulted in significantly lower alignment ($p\ll0.001$), suggesting that this LM produces the most brain-aligned embeddings. 
Trends in the face-selective ROI were similar: MS COCO captions were maximally aligned when embedded with Qwen3, but significantly less aligned than the other caption types embedded with the same LM. Captions generated by Molmo were, again, associated with the highest coefficient estimates when embedded with Qwen3 ($\beta=0.17$, $SE=0.01$). In the place-selective ROI, mixed-effects models highlighted different patterns. All machine-generated captions except the Phi-4-generated ones were significantly \textit{less} ($p<0.001$) brain-aligned than MS COCO captions. However, this seemed to happen only when computing embeddings with Qwen3 Embedding; when using other LMs, estimates associated with the MS COCO captions were among the lowest.

Altogether, these results show that, similarly to what was observed for brain encoding, the \textit{LM} used to embed captions yields more consistent brain-alignment gains than the \textit{caption type}. More specifically, one of the text embedders (Qwen3 or KaLM Embedding) produces the most aligned embeddings in each ROI. Regarding caption types, we observed several significant \textit{LM} $\times$ \textit{caption type} interactions in all ROIs, suggesting that the average length or descriptiveness of captions is not sufficient to determine how brain-aligned they will be.

\begin{figure}[h!]
    \centering
    \begin{subfigure}[t]{0.48\textwidth}
        \centering
        \includegraphics[height=1.7in]{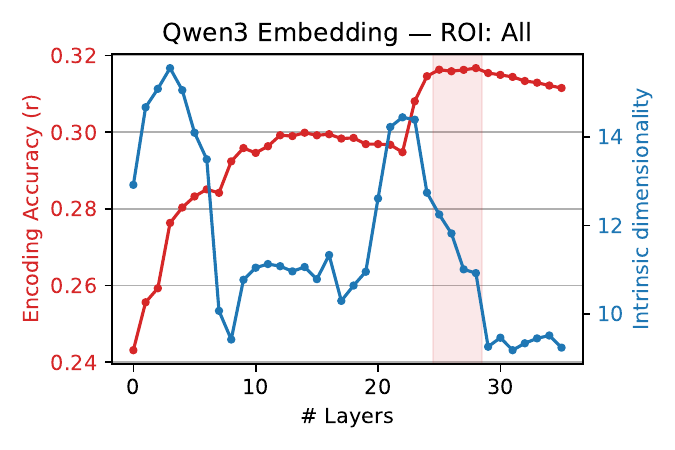}
        % \caption{}
    \end{subfigure}
    \hfill
    \begin{subfigure}[t]{0.48\textwidth}
        \centering
        \includegraphics[height=1.7in]{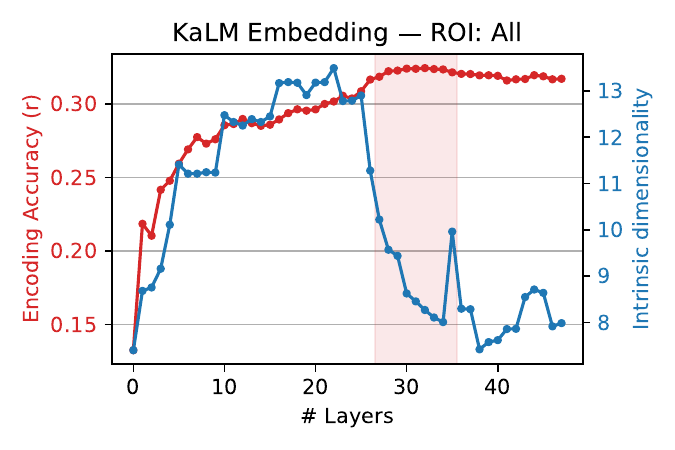}
        % \caption{}
    \end{subfigure}

    \caption{\textbf{Brain encoding performance and intrinsic dimensionality.} On the left y-axis, line charts show the average brain encoding performance at each layer, computed across all voxels from place-, face, and body-selective ROIs. On the right y-axis, intrinsic dimensionality as estimated with the GRIDE method, providing an indication of how compressed information is at each model layer. The original dimensionality of Qwen3 embeddings was $4096$, while that of KaLM embeddings was $3840$. Light-red shading highlights the point where encoding accuracy peaks.}
    \label{fig:compression_analysis}
\end{figure}

% ------------------------------------------------------------
\subsection{Brain encoding peaks after a phase of high intrinsic dimensionality}
\label{sec:layers_encoding}
% ------------------------------------------------------------
To better understand what type of information is responsible for the brain predictivity and alignment of caption embeddings, we examined how these metrics vary throughout LM layers, and how they relate to information compression. While learning representations, LMs compress information by projecting the input text into manifolds having a `true' (\textit{intrinsic}) dimensionality much lower than the actual embedding dimensionality \cite{cheng2023bridging, valeriani2023}, which is constrained by the architecture and fixed across layers. Previous studies on information compression in LMs have shown that peaks in intrinsic dimensionality signal the onset of abstract linguistic information about semantics and syntax  \cite{valeriani2023, razzhigaev-etal-2024-shape, cheng2025emergence}. 

Motivated by these findings, we estimated intrinsic dimensionality using the Generalised Ratios
Intrinsic Dimension Estimator (GRIDE, \cite{denti2022generalized}), a recent non-linear approach (see Methods for additional details).
This method was applied to embeddings by Qwen3 and KaLM, which consistently yielded the highest brain predictivity and alignment; results for the remaining architectures are provided in Supp. Fig. 9. 
In Fig.~\ref{fig:compression_analysis}, we visualise intrinsic dimensionality along with brain encoding performance at each model layer. Brain predictivity was averaged across voxels from all ROIs, as we did not observe substantial differences between them (but see Supp. Fig. 8 for individual ROIs). This analysis shows that brain predictivity peaks immediately after the last high-dimensional phase in both Qwen3 and KaLM Embedding. Given that this phase is thought to signal the emergence of abstract linguistic information \cite{cheng2025emergence}, our findings suggest that meaningfully representing the semantic and syntactic properties of image captions is beneficial for modelling neural activations in high-level visual cortices.

\begin{figure}[]
    \centering

    % ---------- Row 1 ----------
    \begin{subfigure}[t]{0.423\textwidth}
        \centering
        \includegraphics[width=\textwidth]{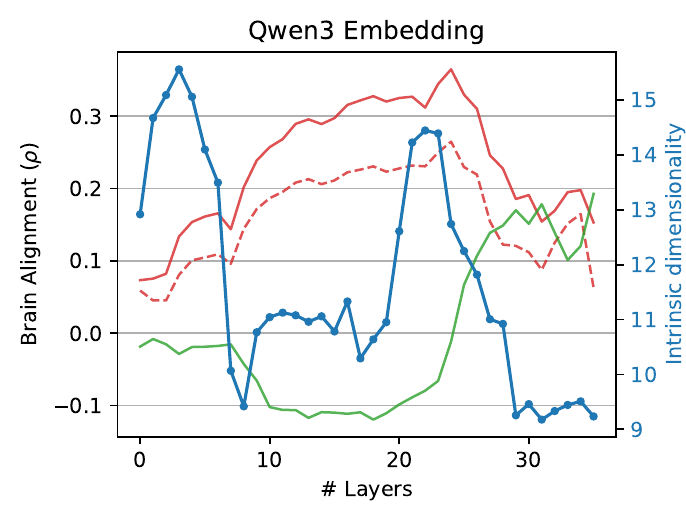}
    \end{subfigure}
    \hfill
    \begin{subfigure}[t]{0.555\textwidth}
        \centering
        \includegraphics[width=\textwidth]{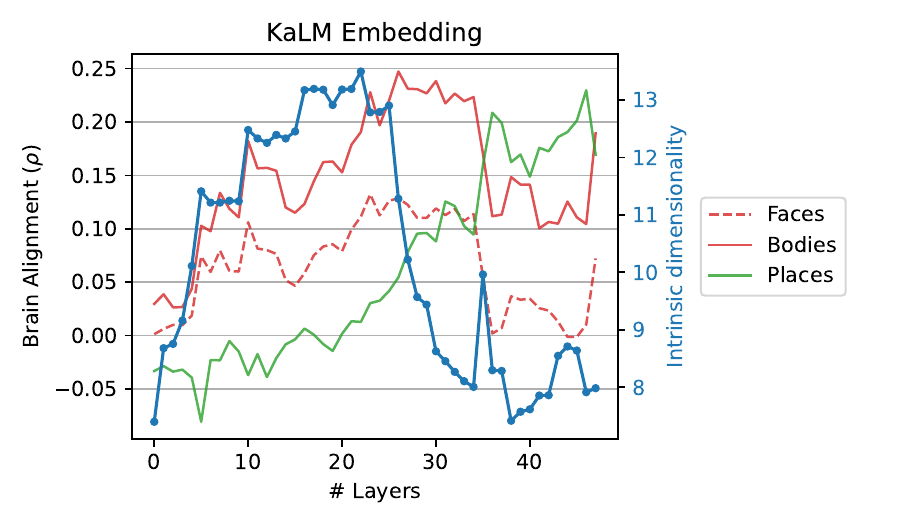}
    \end{subfigure}

    \caption{\textbf{Intrinsic dimensionality and brain alignment.} Layer-wise brain alignment exhibited by Qwen3 Embedding (left-hand panel) and KaLM Embedding (right-hand panel) in the three ROIs (place-, face-, and body-selective), expressed as Spearman correlations ($\rho$) normalised by the noise ceilings. Intrinsic dimensionality estimated with the GRIDE method is displayed in blue in both panels.}
    \label{fig:brain_rsa_layers}
\end{figure}

% ------------------------------------------------------------
\subsection{Alignment to body-/face-selective regions and place-selective ones is dissociated within model layers}
\label{sec:layers_rsa}
% ------------------------------------------------------------

The same layer-wise analysis was conducted for brain alignment; in this case, we report the layer-wise progression separately for each ROI (Fig.~\ref{fig:brain_rsa_layers}), as we did observe meaninfgul differences. Indeed, trajectories of brain alignment through layers show a `double dissociation' between face-/body-selective ROIs and the place-selective region. Alignment with face-/body-selective ROIs increases through the mid layers and diminishes in the latest; conversely, alignment in the place-selective ROI remains low in the early-to-mid layers and increases dramatically in the latest. This pattern can be observed in both Qwen3 and KaLM Embedding. Intriguingly, this suggests that similarity patterns in face-/body-selective ROIs and in place-selective regions rely on diverging information.

Relating these observations to intrinsic dimensionality, Fig.~\ref{fig:brain_rsa_layers} shows that alignment peaks immediately after the last high-dimensional phase in body-/face- selective regions. As for the place-selective ROI, the highest alignment was observed towards the last layers, significantly later than the high-dimensional phase. This could mean that the linguistic information emerging after the high-dimensional phase is useful to model similarity patterns in body-/face-selective ROIs, but not in the place-selective one. 

Finally, Figs.~\ref{fig:compression_analysis} and~\ref{fig:brain_rsa_layers} afford a comparison between layer-wise patterns in brain predictivity and brain alignment. In Qwen3 Embedding, predictivity peaks in the layer interval 26--29, similarly to alignment with place-selective areas (peaking at layer 31), but later than alignment with face- and body-selective ROIs (reaching its maximum at layer 24). In KaLM Embedding, brain predictivity peaks in the layer interval 31--35; this partially overlaps with the maximal alignment with face- and body-selective ROIs (layers 24--34), while alignment to place-selective areas peaks later (layer 46). Altogether, these observations indicate that the overlap between the layers where brain predictivity and alignment reach their maximum is partial, and limited to specific ROIs. This invites caution against assuming that the information relevant to directly model neural activations and the information useful for modelling their similarity patterns reside in the same model layers.

\begin{figure}[h!]
    \centering

    % ---------- Row 1 ----------
    
    \begin{subfigure}[t]{0.65\textwidth}
        \centering
        \includegraphics[width=\textwidth]{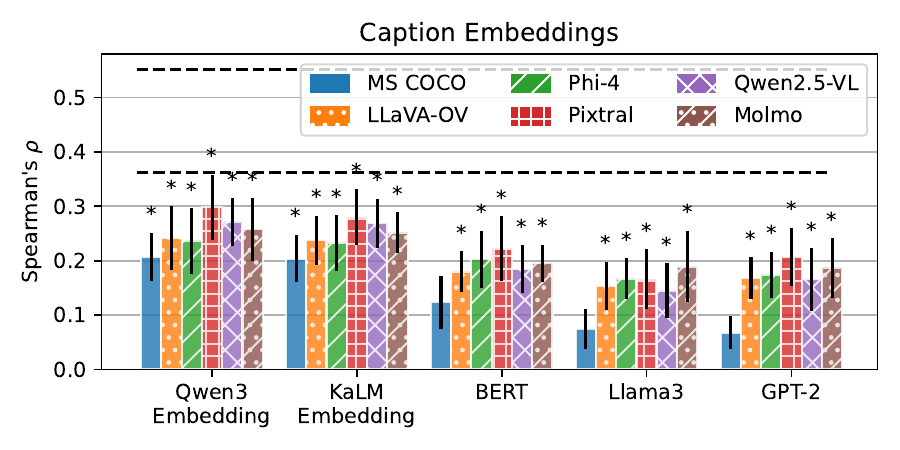}
        \caption{}
    \end{subfigure}
    % \hfill
    \begin{subfigure}[t]{0.32\textwidth}
        \centering
        \includegraphics[width=\textwidth]{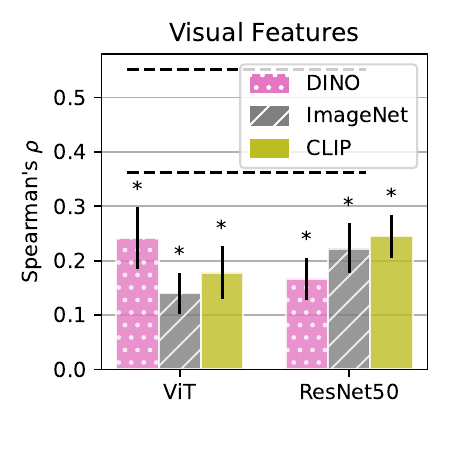}
        \caption{}
    \end{subfigure}

    % ---------- Row 2 ----------
    \begin{subfigure}[t]{0.48\textwidth}
        \centering
        \includegraphics[width=\textwidth]{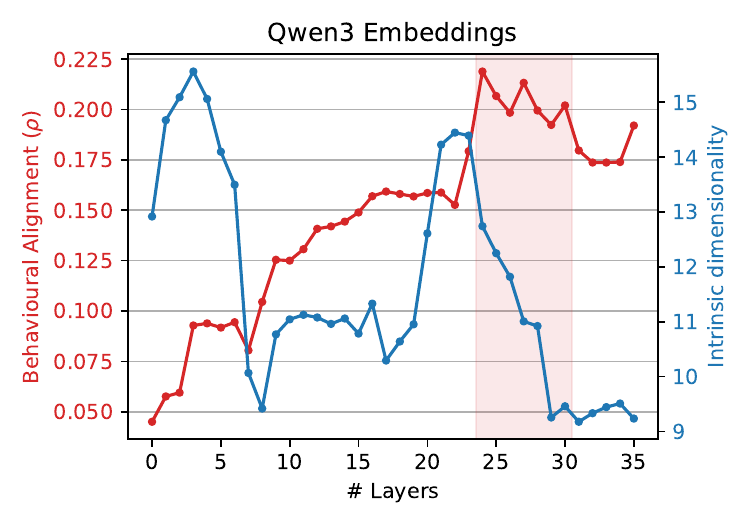}
        \caption{}
    \end{subfigure}
    \hfill
    \begin{subfigure}[t]{0.48\textwidth}
        \centering
        \includegraphics[width=\textwidth]{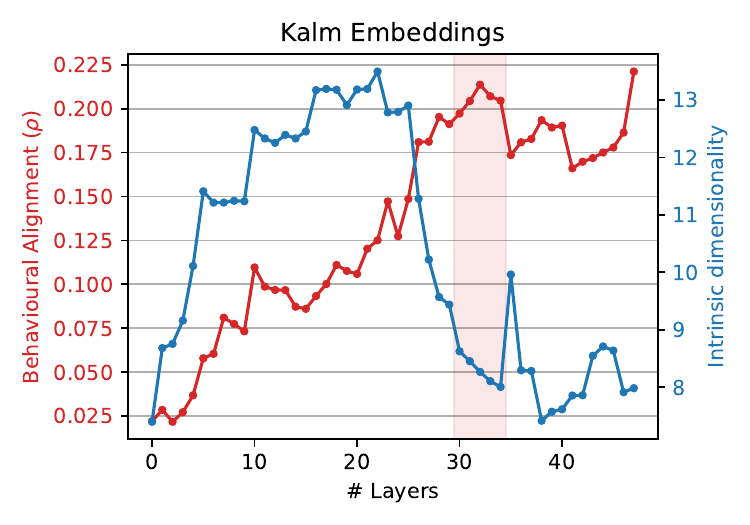}
        \caption{}
    \end{subfigure}

    \caption{\textbf{Behavioural alignment achieved by caption embeddings.} In the top panels, results from representational similarity analysis measuring the alignment (Spearman's $\rho$) between human similarity judgments and caption embeddings (\textbf{a}) or visual features (\textbf{b}). Results are averaged across participants, selecting the most aligned model layer participant-wise. Error bars indicate the standard deviation across participants. Dashed lines represent the upper and lower bounds of the noise ceilings. Asterisks indicate that correlations were statistically significant ($p<0.001$) for all participants, as assessed with permutation tests (1000 permutations). In non-significant cases, correlations were non-significant for at most 2 participants. In the lower panels, the layer-wise behavioural alignment exhibited by Qwen 3 (\textbf{c}) and KaLM Embedding (\textbf{d}) is plotted jointly with the intrinsic dimensionality at each layer, as estimated with the GRIDE algorithm. Light-red shading highlights the point where behavioural alignment is the highest.}
    \label{fig:simj}
\end{figure}

% ------------------------------------------------------------
\subsection{Caption embeddings by text embedders capture human-perceived similarity better than image features}
\label{sec:judgments}
% ------------------------------------------------------------

To gain a comprehensive understanding of the cognitive-modelling potential of caption embeddings, we additionally measured their ability to approximate behavioural data, in the form of image similarity judgments.
Previous studies have found that similarity judgments reflect additional semantic properties not represented in high-level visual areas \cite{groen2018distinct, king2019similarity}, as they are likely informed by cognitive processes beyond perception. For this reason, we expected caption embeddings to compare even more favourably to image features when modelling behavioural responses vs.~neural data.

To test this hypothesis, we analysed a set of image-similarity judgments included in the NSD, but never studied by previous work. These were collected by asking the same participants who were fMRI scanned to spatially arrange 100 images (a subset of the 906 from the fMRI experiment) based on their similarity. We again used RSA to measure the alignment between these human similarity judgments and the caption embeddings/image features considered in the fMRI-data analyses described above (see Fig.~\ref{fig:evaluation_pipeline}). 

The behavioural alignment achieved by both kinds of representations is visualised in Fig.~\ref{fig:simj}. These results show that some of the caption embeddings---those obtained by feeding KaLM and Qwen3 Embedding with Pixtral and Qwen2.5-VL-generated captions---are more behaviour-aligned than all visual features. In addition, Pixtral-generated captions embedded with Qwen3 approach the lower bound of the noise ceilings. These observations indicate that the content of an image, as represented by an embedded caption, can provide an accurate model of human-perceived image similarity, to an even larger extent than image features.

To better understand what---whether the captions themselves or how they are embedded---drives the behavioural alignment of caption embeddings, we again analysed our results with mixed-effects models. The settings and parameters were the same as those reported above for brain alignment.
The reference condition was, again, MS COCO captions embedded with Qwen3, and yielded a coefficient estimate of $\beta=0.21$ ($SE=0.02$, $t=11.22$). Embedding the same MS COCO captions with BERT, GPT-2 or Llama3.1 resulted in statistically significant ($p<0.001$) estimate decreases, while the difference was not significant for KaLM Embedding. Moreover, all other caption types were associated with significantly higher ($p<0.05$) coefficient estimates. This suggests that Qwen3 and KaLM Embedding yield the most behaviour-aligned caption representations, and that MS COCO captions tend to be less behaviour-aligned than machine-generated captions. A few significant interactions indicate that, again, some caption types are more aligned when embedded with specific LMs. For example, the interaction was positive and significant for Phi-4-generated and Molmo-generated captions embedded with GPT-2 (respectively, $\beta=0.28$, $SE=0.02$, $t=3.77$, and $\beta=0.28$, $SE=0.02$, $t=3.35$).  

We next checked how layer-wise behavioural alignment relates to intrinsic dimensionality, focusing on the most aligned architectures: Qwen3 and KaLM Embedding. As shown in Fig.~\ref{fig:simj}, alignment with similarity judgments peaks after the (last) high-dimensional phase, thought to cue the emergence of abstract linguistic information. Intriguingly, maximal behavioural predictivity was observed in the same layers where neural predictivity is at its highest. This suggests overlap between the information useful to predict brain responses and that driving behavioural alignment.

% =============================================================
\section{Discussion}
\label{sec:results}
% =============================================================
% [should be 1200--1500 words, it is now around 1100]

The present study extensively investigates image-caption embeddings as a means to model high-level visual perception. We analysed neural activations by measuring predictivity and alignment, allowing us to study both the feature space defined by embedded captions and its representational geometry. Moreover, we provided the first analysis of the NSD similarity judgments, finding that caption embeddings can achieve higher behavioural alignment than image features extracted from visual models. We also found suggestive evidence that machine-generated image captions yield significant brain predictivity and behavioural alignment, sometimes surpassing the human-annotated captions used in previous work. Below, we discuss each of our findings in detail.

Brain encoding revealed that image descriptions embedded with language models significantly predict neural activations in place-, body-, and face-selective visual areas, in most cases only marginally worse than image features by vision models. This aligns with previous studies, which also documented near-image-feature predictivity for embeddings of MS COCO captions in high-level visual areas \cite{wang2023better, doerig2025high, conwellrethinking}. Altogether, these results fit with the widely-accepted notion that high-level visual areas are selective for specific \textit{semantic} categories \cite{scherf2007visual, de2019factors, peelen2017category}. In this sense, it stands to reason that caption embeddings---presumably providing accurate semantic representations of an image---significantly predict their neural activations. 

Extending previous work, we investigated whether brain predictivity is sensitive to the captions themselves or to the language models used to embed them. Regarding the first factor, a recurrent finding was that the human-annotated captions from the MS COCO dataset---the only ones analysed in previous work---tend to be less brain-predictive and brain-aligned than the other, machine-generated caption types. However, one important methodological difference between our study and previous work is that we randomly sampled one MS COCO caption per image, while existing studies averaged the embeddings of the five captions available in MS COCO for each image. Upon re-computing brain predictivity on averaged embeddings (see Supp. Fig. 7), we observed increased predictivity, comparable to that achieved by non-averaged machine-generated captions. This suggests that a single, `good' image description is sufficient to predict neural activations as effectively as less `good' descriptions do only when averaged across multiple instances. 

What makes a caption `good' or `bad' from a neural and cognitive modelling perspective, then? Previous work highlighted brevity and lack of detail as potential limitations with respect to brain predictivity \cite{doerig2025high}. While we did observe higher brain predictivity and behavioural alignment for longer caption types, this trend did not hold as consistently when measuring brain alignment. We therefore hypothesise that an additional limitation of MS COCO captions concerns their fluency, which was substantially lower than that of the machine-generated captions. 

Documenting higher fluency for \textit{machine}-generated captions may appear surprising, or even inflated by the use of perplexity as a measure for fluency---language models are known to generate more predictable text than humans \cite{liao2023differentiating}. However, human annotators, unlike machines, are subject to cognitive constraints, causing their accuracy to decrease over time due, e.g., to fatigue \cite{hopstaken2015multifaceted} or low motivation \cite{rogstadius2011assessment}. The consequences of these constraints become even more prominent when collecting annotations outside a lab through an online platform, as was the case for MS COCO captions. In light of these considerations, MS COCO captions can be regarded as noisy annotations,
explaining why averaging over multiple instances is necessary to achieve brain predictivity comparable to machine-generated captions. On the other hand, hallucinations (mentions of entities that are not really present in an image) could be a source of noise for \textit{machine}-generated captions \cite{rohrbach-etal-2018-object, Biten_2022_WACV}. While they did not seem to be severe in our setting, more research is needed to systematically evaluate their impact on brain predictivity and alignment.

Concerning the language models used to embed the captions, our experiments revealed systematic trends. By testing an early model pretrained with masked language modelling, two autoregressive models pretrained with next-word prediction, and two text embedders extensively fine-tuned on semantic tasks, we established that the latter exhibit the highest modelling performance across evaluation methods (RSA  vs.~brain encoding) and data modalities (neural activations vs.~behavioural judgments). 
Due to their extensive training on supervised semantic tasks, recent text embedders are better suited than autoregressive LLMs to represent textual meaning, as assessed with text-embedding benchmarks commonly used by NLP practitioners \cite[e.g.,][]{muennighoff-etal-2023-mteb, enevoldsen2025mmteb}. 
% , as they are extensively trained on supervised semantic tasks. In addition, autoregressive LLM training, using unidirectional attention, privileges next-word prediction over meaning representation \cite{zhang-etal-2025-diffusion, su-etal-2025-training, k-etal-2025-large, behnamghader2024llmvec}. 
Remarkably, we found that this superiority also holds in a rather different domain: neural and cognitive modelling of visual perception. This suggests that well-representing the meaning of a caption matters for brain and behavioural alignment, echoing previous findings showing that LLMs' brain predictivity in high-level visual areas is mainly driven by the semantics of the words appearing within captions \cite{doerig2025high, conwellrethinking}.

% Given that our use-case involved embedding full sentences, it was expected that embeddings produced by text embedders would be superior from a technical point of view. Nevertheless, it is non-trivial that this technical superiority transfers to brain predictivity---a domain rather different from the text-embedding tasks typically considered by NLP practitioners. 

We additionally analysed the evolution of brain predictivity and alignment throughout LM layers. We found that both peak in mid-late layers (although not at the exact same point), as opposed to the very early or late ones. Interestingly, this aligns with well-documented findings from previous works where LM embeddings were used to model brain responses to linguistic stimuli \cite{schwartz2019, schrimpf2021neural, caucheteux2022brains, goldstein2025temporal}. We then related layer-wise predictivity patterns to information compression, studied layer-wise by computing intrinsic dimensionality. This analysis showed that both brain predictivity and alignment with similarity judgments tend to peak immediately after a high-dimensional phase, thought to subsume abstract linguistic processing \cite{valeriani2023, cheng2025emergence}. Intriguingly, our findings align with preliminary evidence in the language domain, suggesting that the brain-encoding properties of language and speech models are attributable to meaning abstraction, signalled by a peak in intrinsic dimensionality \cite{cheng2026abstraction}, as opposed to next-word prediction as previously argued \cite{schrimpf2021neural, goldstein2022shared, caucheteux2023evidence}. 

An interesting finding emerged when studying brain alignment throughout layers was that the information relevant for modelling similarity patterns in face-/body-selective ROIs appears to be dissociated from that relevant for the place-selective ROI. Indeed, the layers where alignment to face-/body-selective areas was maximal exhibited extremely low alignment to the place-selective ROI, and vice versa. A related effect was observed in previous work, finding representational patterns to differ significantly between the fusiform face area (FFA) and the occipital place area (OPA) \cite{king2019similarity}; the same study also reported that image representations extracted from deep neural networks (DNNs) exhibited significant alignment with the FFA at layer depths where the correlation with the OPA was negative. As for the similarities we observed between alignment patterns in face- and body-selective areas, they echo preliminary evidence suggesting that these two regions may be performing shared computations, as demonstrated by the fact that DNN units sensitive to both faces and bodies predict fMRI activations in place- and body-selective ROIs better than units sensitive to solely faces or bodies \cite{van2026face}. 

One last notable finding from our study concerns the NSD similarity judgments, which, to the best of our knowledge, we analysed for the first time. When computing behavioural alignment, caption embeddings sometimes yielded higher correlations than image features. This result aligns with findings from one previous study, which also documented comparable behavioural alignment for embedded image captions and visual features \cite{marjieh2023words}. 
More broadly, these observations fit with the idea that perception and language are deeply related and mutually influence each other \cite{lupyan2012linguistically, lupyan2015words}, although their complex interplay is still far from being fully understood \cite{simanova2016linguistic}.

Altogether, our study shows that image captions are an effective means to model not only neural activations in high-level visual areas but also perceived similarities between images. In addition, our findings indicate that, while differences between caption types may be sensitive to the evaluation protocol (encoding vs. alignment), brain ROIs, and data modalities (fMRI vs. behavioural), embedding them with a model that effectively represents the sentence-level meaning results in consistent neural and cognitive modelling gains. We hope our contribution inspires future computational work on modelling visual perception, as well as efforts to better understand its relations with linguistic knowledge.    

% =============================================================
\section{Methods}
\label{sec:methods}
% =============================================================
\paragraph{Stimuli} The 906 images shown to participants in the fMRI experiment were selected from the MS COCO dataset \cite{lin2014microsoft} and are publicly available with different types of annotations, including captions written by crowdworkers. For the fMRI experiment, images were preprocessed by applying some transformations (e.g., cropping and downsampling; see \cite{allen2022massive} for additional details). We employed the same preprocessing before feeding the images to vision models and vision-language models.

\paragraph{fMRI responses} The fMRI responses we analysed are from the Natural Scenes Dataset (NSD, \cite{allen2022massive}), a large collection of high-resolution (7T) brain recordings. 
We consider a subset of the brain responses, i.e., those elicited by the `Special 1000' set, viewed at least once by all eight participants. 
Subjects took part in multiple scanning sessions and were instructed to perform a continuous recognition task where they had to determine whether they saw each image at any previous point in the experiment. The experimental design consisted in presenting each image three times throughout all sessions. Unfortunately, not all participants were able to complete all the scanning sessions, resulting in a final set of 906 images viewed at least once by all participants. 

We used the fMRI recordings as preprocessed by the dataset curators. More specifically, we focused on the voxel-wise beta estimates of the fMRI response amplitude computed with a three-component GLM approach (`betas\_fithrf\_GLMdenoise\_RR' coefficients) in the subjects' native surface. 
These beta coefficients are the neural activations we aimed to predict. The regions of interest we focused on are \textit{Places}, \textit{Bodies}, and \textit{Faces}. These were functionally localised with dedicated tasks, completed by all participants before the main experiment. We refer the reader to the original paper for further details on data collection, preprocessing, and localisation of ROIs.

\paragraph{Behavioural similarity judgments} The curators of the NSD collected image similarity annotations from the same subjects who took part in the fMRI experiments for 100 images selected from the Special 1000 set. These annotations were collected through a multiple-arrangements task \cite{kriegeskorte2012inverse} where participants were asked to drag and drop images on a circular arena based on their similarity. 

\paragraph{Image captions}
We analysed six different caption sets, including those from MS COCO and five sets of machine-generated captions. Given that five different MS COCO captions are available for each image, we randomly sampled one. As for the machine-generated captions, we obtained them from five recent vision-language models: Molmo \cite{deitke2024molmo}, LLaVA OneVision \cite{li2025llavaonevision}, Qwen2.5-VL \cite{bai2025qwen2}, Phi-4 \cite{abouelenin2025phi}, and Pixtral Large \cite{pixtral}. All models were accessed through the `HuggingFace Transformers' Python library \cite{wolf2019huggingface}; the specific model IDs are included in our public GitHub repository. Models were prompted by asking for short (max. two sentences) and factual image descriptions (see Supp. Tab. 1 for the model-specific prompts). This length constraint was introduced to avoid overly verbose descriptions providing non-relevant information. A manual inspection of the captions revealed that they are mostly correct (i.e., describing content present in the image), containing only minor mistakes (e.g., mentioning the wrong number of objects) or hallucinations.  

\paragraph{Caption metrics}
The metrics we computed to characterise the differences between caption types include perplexity, visualness, and some variants of lexical density. Given a sequence made up of $N$ words $w_1$, $w_2$, ..., $w_N$ the perplexity of that sequence is formally defined as:
\begin{equation}
    exp\left(-\frac{1}{N}\sum_{i=1}^N\ln \left[P(w_i|w_{i-1}, w_{i-2},...,w_{1})\right]\right),
\end{equation}

where $P(w_i|w_{i-1},...,w_{1})$ is the predicted probability of the $i^{th}$ word. We computed perplexity with a custom Python script, using the probabilities output by the Ministral3 language model \cite{liu2026ministral}, accessed via the `HuggingFace Transformers' library \cite{wolf2019huggingface}.

The visualness score for a caption was computed by leveraging the Lancaster Sensorimotor Norms \cite{lynott2020lancaster}, which collect human ratings on sensory-motor dimensions for $40,000$ English words. We considered the averaged ratings for the visual modality, ranging from 0 (`not experienced at all with vision') to 5 (`experienced greatly with vision'). For each caption, visualness was computed by summing the vision scores of its lemmatised constituent words. We were able to find a match for 85\% of the \textit{unique} lemmatised words, considering captions from all images and of all types. Words for which no match was detected were discarded.

Finally, lexical density ($LD$) was computed as the percentage of content words within a caption, defined as:

\begin{equation}
    LD = \frac{Number\,of\,Content\,Words}{Total\,Number\,of\,Words} \times 100
\end{equation} 

When computing `overall' lexical density, we considered nouns, adjectives, adverbs, and verbs. Additionally, we computed the density of nouns, adjectives and verbs separately. Captions were POS-tagged (classified according to their part of speech) with the `spaCy' Python package \cite{spacy}.

\paragraph{Caption embeddings} Caption embeddings were computed with five different language models: BERT \cite{devlin-etal-2019-bert}, GPT-2 \cite{radford2019language}, Llama3.1 \cite{grattafiori2024llama}, Qwen3 Embedding \cite{zhang2025qwen3}, and KaLM Embedding \cite{zhao2025KaLM}. BERT is an early transformer-based bidirectional model pretrained in an unsupervised fashion with masked language modelling and next-sentence prediction. GPT-2 is a unidirectional model pretrained on vast amounts of web-scraped texts with next-word prediction. Llama3.1 is also a unidirectional model, pretrained with next-word prediction and then post-trained on tasks aimed at increasing its abilities to follow instructions, align with human output preferences, and perform specific tasks, such as coding or reasoning. 

Qwen3 Embeddings is a unidirectional language transformer initialised with weights from the Qwen3 foundation model \cite{yang2025qwen3} and then trained in multiple steps with a contrastive loss aimed at maximising the similarity between a document and a query. The training stages involve large-scale, weakly-supervised training on noisy datasets, followed by fine-tuning on high-quality datasets and model merging. KaLM Embedding was initialised with weights from Gemma3 \cite{gemmateam2025gemma3technicalreport} and then trained with a contrastive loss, enabling bidirectional attention. Similar to Qwen3 Embedding, the training pipeline involved large-scale pre-training on weakly supervised datasets, fine-tuning on high-quality, supervised datasets, and contrastive distillation on fine-grained soft signals. For both text embedders, the semantic tasks considered for training involved semantic text similarity, text classification, and retrieval.   

All these models were accessed via the HuggingFace Python library \cite{wolf2019huggingface}; the exact model IDs and Python scripts used to compute the embeddings are provided in our public GitHub repository. Embeddings extracted with BERT correspond to the hidden states of the \texttt{CLS} token, i.e., the one storing a `sentence-level' representation. When computing embeddings with GPT-2 and Llama3.1, we considered the hidden states corresponding to the last token of the sequence, a common choice when extracting sentence embeddings from generative language models \cite[e.g.,][]{tuckute2024driving, hosseini2024artificial}. As for Qwen3 Embedding and KaLM Embedding, we extracted sentence embeddings following the instructions in the model cards, as extracting sentence-/document-level embeddings is the intended use case for these models.

\paragraph{Image features}
Image features were computed by feeding vision models with the 906 images viewed by participants in the fMRI experiment, using the same preprocessing. We focused on the convolutional neural network ResNet-50 \cite{he2016deep} and the vision transformer ViT \cite{dosovitskiyimage}, which have been extensively studied by existing NeuroAI work \cite[e.g.,][]{wang2023better, conwell2024large, bartnik2025}, and considered implementations varying in the amount of language supervision provided during training. More specifically, we tested both ViT and ResNet-50 in their original implementations trained for image classification on ImageNet \cite{deng2009imagenet} labels, and in the CLIP implementation \cite{radford2021learning}, trained with a contrastive loss maximising the similarity between images and their captions. We also considered implementations trained purely visually in a self-supervised fashion. These were DINOv2 \cite{oquab2024dinov} for ViT and DINO \cite{caron2021emerging} for ResNet-50 (ResNet-50 is not included in the more recent DINOv2 framework). We computed image features (\textit{hidden states}) from each layer of the ViTs. As for ResNet-50, we extracted representations at five different depths in the network (\texttt{stem}, \texttt{conv1}, \texttt{conv2}, \texttt{conv3}, \texttt{conv4}, \texttt{attnpool}).

\paragraph{Voxel-wise encoding models}
To assess the brain predictivity of the different caption embeddings and image features, we predicted the brain activations in each voxel from the three ROIs (\textit{Faces}, \textit{Places}, and \textit{Bodies}) using \textit{voxel-wise encoding models} \cite{naselaris2011encoding, dupre2025voxelwise}. Model embeddings from a specific layer, denoted by $X \in {R}^{n \times d}$ (with $n$ stimuli and $d$ features), 
were first downsampled with principal component analysis (PCA), adaptively selecting the number of components $k$ such that $99\%$ of the original variance was captured. This yielded the reduced embeddings $Z \in {R}^{n \times k}$. Next, these downsampled embeddings $Z$ were used as predictors in a Ridge 
regression model:
\[
\hat{y}_v = Z\boldsymbol{\beta}_v, 
\quad 
\boldsymbol{\beta}_v = \arg\min_{\boldsymbol{\beta}} 
\left\{ \lVert y_v - Z\boldsymbol{\beta} \rVert_2^2 
+ \alpha \lVert \boldsymbol{\beta} \rVert_2^2 \right\},
\]
where $y_v \in {R}^{n}$ represents the measured responses of voxel $v$. 
Note that each voxel $v$ was predicted independently.

Ridge regression was performed in a $5$-fold cross-validation framework. Within each fold, the optimal value of the regularisation parameter $\alpha$ was determined through leave-one-out cross-validation, choosing among $20$ logarithmically spaced values. Ridge regressions were fit using custom Python scripts. To quantify the accuracy of the predicted activations $\hat{y}_v$, we computed Pearson's correlation coefficient $r( \hat{y}_v, v)$, where $v$ are the observed activations for the test set of each fold. In the paper, we reported correlations averaged across test sets of all folds. 
% These correlations were computed through the \texttt{Scipy} Python package. 

\paragraph{Statistical significance of voxel-wise predictions} To assess the statistical significance of each voxel-wise prediction, we conducted permutation tests where we retrained the encoding models as described above, randomly shuffling the labels of the predicted brain responses in the training set. We considered 1000 permutations, allowing us to derive a distribution of correlation values $r$ for each voxel. We then computed a $p$-value as the proportion of $r$ values greater than, or equal to, the observed $r$ value (without any shuffling). For computational efficiency, we did not re-determine the optimal value of $\alpha$ at each permutation, but used the best value as pre-selected with the correct (non-shuffled) labels. Finally, we applied a Benjamini-Hochberg (or false discovery rate) \cite{benjamini1995controlling} correction for multiple comparisons using the `Scipy'  \cite{2020SciPy-NMeth} Python package, with the number of comparisons corresponding to the number of voxels in each ROI. 

\paragraph{Linear mixed-effects models} To systematically determine whether caption type and LM significantly influence brain predictivity, brain alignment and behavioural alignment, we fit linear mixed-effects models using the R packages `lme4' \cite{lmer2015} and `lmerTest' \cite{lmertest}. When analysing brain results, we fit an independent model for each ROI (\textit{Faces}, \textit{Places}, \textit{Bodies}). We included random effects for each participant in all cases and proceeded to fit nested models. We started with a model including `language model' as the only fixed effect (\texttt{y $\sim$ language\_model + (1 | participant)}), then included `caption type' as an additional fixed effect (\texttt{y $\sim$ language\_model + caption\_type + (1 | participant)}), and finally an interaction between the two (\texttt{y $\sim$ language\_model * caption\_type + (1 | participant)}). Each time we added new parameters, we used a likelihood ratio test to determine if the improvements in the fit justified the increased model complexity. In all cases, we reported results for the most complex model that provided a statistically significant improvement ($p\ll0.001$) over the nested ones. 

\paragraph{Representational similarity analysis}
To quantify the brain and behavioural alignment of model representations (caption embeddings and image features), we used Representational Similarity Analysis (RSA,  \cite{kriegeskorte2008representational}). This method quantifies alignment as the Spearman correlation ($\rho$) between representational dissimilarity matrices (RDMs) storing pairwise distances between stimulus representations. In the case of model embeddings (both caption embeddings and image features) and brain responses, these pairwise distances were computed as cosine distances ($1-cosine\,similarity$). As for the similarity judgments, the RDMs derived from the Multiple Arrangements Task were made publicly available by the curators of the NSD. The final Spearman correlation was computed on the vectorised off-diagonals of the RDMs.
RSA was performed separately for each participant and model layer. Pairwise cosine distances and Spearman correlations were computed with the Python packages `Scipy' \cite{2020SciPy-NMeth} and `Scikit-learn' \cite{scikit-learn}. To determine if Spearman correlations were statistically significant, we performed permutation tests ($n_{permutations}=1000$) where the stimulus labels were shuffled before computing the RDMs. This allowed us to obtain a $p$-value by calculating the proportion of permutations yielding correlations lower than or equal to the observed one.   

We additionally computed noise ceilings, which provide an indication of the maximum representational alignment that can be expected given the inter-participant differences, in both neural activations and behavioural judgments. These noise ceilings were obtained by computing the mean Spearman correlation between each participant's RDM and the average (across participants) RDM, calculated either including the target participant (\textit{upper bound}) or with a leave-one-out approach (\textit{lower bound}) \cite{nili2014toolbox}.   

% \paragraph{Principal component analysis} When using principal component analysis to analyse information compression in LM layers, we relied on the implementation provided within the `Scikit-learn' \cite{scikit-learn} Python package. We computed the number of principal components necessary to explain 99\% of the variance from all caption types.

\paragraph{Intrinsic dimensionality} To estimate the intrinsic dimensionality at each layer of the language models, we used the generalised ratios intrinsic dimension estimator (GRIDE, \cite{denti2022generalized}). This is a nearest-neighbour (NN) method, i.e., it assumes that data points close to each other are uniformly drawn from $d$-dimensional hyperspheres, and it estimates intrinsic dimensionality as a function of the average of the distances among the sample points and their respective $k$-th NN, with $k$ being the number of NNs considered. 

More formally, consider the random variable $\dot{\mu} = \mu_{i,n_1,n_2} = \frac{r_{i,n_2}}{r_{i,n_1}}$, where $r_{i,l}$ is the distance between observation $i$ and its $l$-th NN and $1\le n_1 < n_2$ are integers. The density function of this random variable is:
\begin{equation}
    f_{\mu_{i, n_1, n_2}(\dot{\mu})} = \frac{d(\dot{\mu}^d-1)^{n_2-n_1-1}}{\dot{\mu}^{(n_2-1)d+1}B(n_2-n_1,n_1)}, \dot{\mu}>1,
\end{equation}

\noindent where $B(\cdot, \cdot)$ denotes the Beta function and $d$ is the intrinsic dimensionality. Following a maximum-likelihood approach and setting $n_2=n_1+1$, it is possible to derive the following estimator: 

\begin{equation}
    \hat{d}_L = \frac{n(L-1)-1}{\sum^n_{i=1}\sum^L_{l=2}(l-1)log(\mu_{i,l})},
\end{equation}
\noindent where $L$ indicates the maximum NN rank considered. Correctly identifying the number of intrinsic dimensions, therefore, requires computing estimates considering different neighbourhood sizes. 

When conducting our experiments, we estimated intrinsic dimensionality with the GRIDE algorithm as implemented in the `Dadapy' \cite{dadapy} Python package. We set the maximum nearest neighbour rank ($L$) to 5436 (906 images $\times$ 6 caption types), resulting in 12 estimates ($log_2(L)$, as the NN order is doubled at each estimate).
 
\bibliography{sn-bibliography}  

\section*{Acknowledgements}
This project originated during A. B.’s research visit to M.-F. M. and her LIIR lab at KU Leuven. This marked the beginning of a fruitful research collaboration and created the opportunity for stimulating discussions with Helena Balabin, whom we warmly thank for the support and insights provided at the early stages of this project.  Our gratitude also goes to the members of the Amsterdam Dialogue Modelling Group and Mulini Lab for the helpful feedback provided at different stages of the project. A heartfelt thank you to Iris Groen and Ece Takmaz for insightful discussions that helped shape the experimental design of the study. We additionally thank Davide Marcantonio for providing valuable feedback on an earlier version of the manuscript.

\section*{Author Contributions}
A. B. designed the study, conducted all neural data experiments, visualised the results, performed the primary data analyses, and drafted the manuscript. I. K. designed and conducted the behavioural experiment and contributed to manuscript revision. R. F. and M.-F. M. supervised the project, provided substantial input on experimental design, data analysis, interpretation of the results, and manuscript writing.

\end{document}